\documentclass[conference, letterpaper]{IEEEtran}
\ifCLASSINFOpdf
\else
\fi
\ifCLASSINFOpdf
   \usepackage[pdftex]{graphicx}
\else
\fi

\usepackage[misc,geometry]{ifsym} 
\usepackage{comment}
\usepackage{multirow}
\usepackage{amssymb}
\usepackage{array}
\usepackage{tabularx}
\usepackage[linesnumbered,algoruled,boxed,lined]{algorithm2e}
\SetKwInOut{Parameter}{parameter}
\usepackage{subfig}
\graphicspath{{photos/}}
\usepackage[cmex10]{amsmath}
\usepackage{color}
\usepackage{fancyhdr}

\renewcommand{\thispagestyle}[2]{}

\fancypagestyle{plain}{
        \fancyhead{}
        \fancyhead[C]{first page center header}
        \fancyfoot{}
        \fancyfoot[C]{first page center footer}
}
\begin{document}

%

\title{Search Space of Adversarial Perturbations against Image Filters}

\author{\IEEEauthorblockN{Dang Duy Thang}
\IEEEauthorblockA{Institute of Information Security\\ Yokohama, Japan\\
Email: dgs174101@iisec.ac.jp}
\and
\IEEEauthorblockN{Toshihiro Matsui}
\IEEEauthorblockA{Institute of Information Security\\ Yokohama, Japan\\
	Email: matsui@iisec.ac.jp}}


%


\maketitle

\begin{abstract}
The superiority of deep learning performance is threatened by safety issues for itself. Recent findings have shown that deep learning systems are very weak to adversarial examples, an attack form that was altered by the attacker's intent to deceive the deep learning system. There are many proposed defensive methods to protect deep learning systems against adversarial examples. However, there is still lack of principal strategies to deceive those defensive methods. Any time a particular countermeasure is proposed, a new powerful adversarial attack will be invented to deceive that countermeasure. In this study, we focus on investigating the ability to create adversarial patterns in search space against defensive methods that use image filters. Experimental results conducted on the ImageNet dataset with image classification tasks showed the correlation between the search space of adversarial perturbation and filters. These findings open a new direction for building stronger offensive methods towards deep learning systems.
\end{abstract}



%
\IEEEpeerreviewmaketitle

\section{Introduction}
Over the past decade, there has been the rise of deep learning in many tasks such as computer vision~\cite{ioannidou2017deep}, automatic driving~\cite{watzenig2016automated}, natural language processing~\cite{ray2018rise}, and so on . Deep learning models are designed based on an assumption of inputs and outputs distribution being benign. This leads to when training deep learning models, we only focus to fine-tune the weights, parameters or the number of nodes and hidden layers while setting aside the validity of data. This has created a security issue against deep learning systems. Szegedy et al.~\cite{szegedy2013intriguing} explored that deep neural networks are at risk of attacks from adversarial example attacks. Afterward, many research work on technologies that delude AI models has gradually become a hot spot, and researchers have continued to propose new methods of attack and defense. Adversarial attacks have been regularly adapted in both research and commerce. In the computer vision area, many adversarial attacks are proposed in image classification~\cite{43405, papernot2016limitations, carlini2017, chen2017zoo}, and object detection~\cite{xiao2018characterizing}. 
There are also many researches work on the adversarial example in text~\cite{miyato2016adversarial, ebrahimi2018hotflip, gao2018black, li2019textbugger}. 
In the physical world attack, Kurakin et al.~\cite{kurakin2016adversarial} first exposed that hazards of adversarial examples. They use an application of Tensor-Flow Camera Demo to capture original images. After that, they use Google Inception V3~\cite{szegedy2016rethinking} for classifying those images. The implementation has been shown that a large portion of the image has been misclassified even when observed via the camera lens. Eykholt Kevin et al.~\cite{eykholt2017robust} invented a new method based on~\cite{carlini2017} and~\cite{liu2016delving} to create robust adversarial perturbation in the real world. They indicated variation in view angles, distance, and resolution are almost defeated by the robust adversarial examples in physical settings. The proposed algorithm used a term as \(RP_2\) for Robust Physical Perturbations, which was used to craft adversarial examples for road sign recognition systems that perform a high deceiving rate in an efficient setting. And many physical adversarial attacks are proposed in face recognition~\cite{sharif2016accessorize}, machine vision~\cite{melis2017deep}, and road sign recognition~\cite{eykholt2018robust}.
In the cyberspace security field, there are adversarial attacks in cloud service~\cite{papernot2017practical}, malware detection~\cite{grosse2016adversarial, anderson2016deepdga}, and network intrusion detection~\cite{huang2018adversarial}. Besides the attack methods, many defensive approaches have been proposed and they can be branched into four main categories include adversarial training, denoising, transformation and compression. Szegedy et al.~\cite{szegedy2013intriguing} used adversarial examples to train an AI model with the ground truth labels, and it made that model more robust. Goodfellow et al.~\cite{43405} also used the adversarial training strategy to improve the classification rate on adversarial examples with the MNIST dataset. Tramèr et al.~\cite{tramer2017ensemble} combined the adversarial examples created from many different AI models to increase the robustness of those models.~\cite{DBLP:journals/corr/abs-1901-03037, 8958446} proposed new methods based on the image transformation to reduce the misclassification rate of an AI model.~\cite{thang2019automated, dangduythang2019label} assumed almost adversarial examples are created in the high-frequency domain and they proposed the new method based on image filters to remove the adversarial perturbations. Das et al.~\cite{das2017keeping} introduced a defensive method based on JPEG compression to deceive FGSM~\cite{43405} and DeepFool~\cite{moosavi2016deepfool} attacks. However, newer adversarial attacks such as Carlini\&Wagner attacks~\cite{carlini2017} overcame these compression defensive strategies.

\textbf{Our Contributions.}
In this work, we investigate the search space of adversarial perturbation. A challenge in the process of understanding the effects of adversarial noises is very limited so far. How to determine the available space of adversarial noises is very important. Understanding and identifying this space will help us develop better protection systems for deep learning against adversarial examples. 

We describe our main contributions of this research as below:
\begin{itemize}
	\item We have recapped the numerous adversarial defensive and attack methods. Moreover, we have provided a perceptive review of these current methods. 
	\item We discovered the close relationship between search space of adversarial perturbation and image filters.  
	\item Our research opens up a new perspective on creating stronger and more effective attacks on deep learning systems.
\end{itemize}

\textbf{Paper outlines.}
The remainder of our paper is described as follows. Section~\ref{background} introduces the literature review and the background of adversarial examples. Section~\ref{proposedmethod} describes our approach on search space of adversarial examples, and Section~\ref{experiements} demonstrate our implementation and evaluation results. Section~\ref{conclusion} summaries our work.
\section{Literature review and Background}\label{background}

\subsection{Literature review} In this work, we focus on the relation between feasible space of adversarial perturbation and defensive methods based on frequency domain. So we make a literature review on these defensive methods in this section. Eliminating the adversarial features and retaking the classification rate have been considered in many works. Xu et al.~\cite{DBLP:conf/ndss/Xu0Q18} proposed a new defensive approach by using the feature squeezing strategies to remove the adversarial features. There are two key ideas in~\cite{DBLP:conf/ndss/Xu0Q18}. The first one considered the bit depth in an input image. By increasing or reducing the bit depth of image, the method removed some adversarial features. The second one used the median filter to defeat the adversarial features. However,~\cite{DBLP:conf/ndss/Xu0Q18} required a range of thresholds to separate between adversarial and legitimate features. So the selection of a relevant threshold for a specific dataset or setting is a nontrivial task and it is heuristic. 
Dang et al.~\cite{thang2019automated} proposed a detection system for automatically identifying adversarial examples with the image filters (Gaussian, Median filter). The system doesn't require to setup any threshold for distinguishing adversarial and benign images. However, there is unclear how the system is able to suffer the stronger and new adversarial attacks. Our paper shows that Gaussian blurring only works well on the small adversarial perturbation, and it is futile to larger and stronger adversarial perturbation. 
\subsection{Background}
\subsubsection{Convolutional  Neural Networks}
Convolutional Neural Networks (CNNs) are designed to learn the important features from the training dataset to match them with the given labels. CNNs are used in many areas~\cite{ioannidou2017deep, ray2018rise} and provided open-source ~\cite{szegedy2016rethinking}. CNNs include multilayers with many operations to process signals from a lower layer to a higher layer in hierarchy architecture. In this research, we emphasize in image classification task so we only cover the brief fundamentals in this area. In an image classification task, CNNs process an input data \(x\) and try to figure out the best matching output label \(y\) from a set of labels \(Y\). The structure of a CNN can be described as shown in table~\ref{tab:google-inception-v3}. The layers are described in a top-down order from input to output. We can see for this CNN network, the input is a color image of size 299\(\times \)299. The first layer is a convolutional layer whose kernel size is 3\(\times \)3 with a stride of 2. The next convolutional layers also use the same kernel size with a difference with the number of kernels as well as stride. In an inception network, it appears layers called inception layers. These inception layers are different from convolutional layers in that they combine several different kernel sizes at once to extract more important features. The inception layer can also be called inception filters. The last adjacent layer is the logits layer before the softmax function is implemented to calculate the probability for each output label corresponding to the input.
\begin{table}[h]
	\caption{Google Inception Architecture~\cite{szegedy2016rethinking}}
	\label{tab:google-inception-v3}
	\hbox to\hsize{\hfil
		\begin{tabular}{l|>{\centering\arraybackslash}m{2cm}|l}\hline\hline
			layer & patch/stride or note & input \\ \hline \hline
			conv & 3 $\times$ 3/2 & 299 $\times$ 299 $\times$ 3 \\ 
			conv & 3 $\times$ 3/1 & 149 $\times$ 149 $\times$ 32 \\ 
			conv padded &  3 $\times$ 3/1 & 147 $\times$ 147 $\times$ 32 \\
			pool &  3 $\times$ 3/2 & 147 $\times$ 147 $\times$ 64 \\
			conv &  3 $\times$ 3/1 & 73 $\times$ 73 $\times$ 64 \\
			conv &  3 $\times$ 3/2 & 71 $\times$ 71 $\times$ 80 \\
			conv &  3 $\times$ 3/1 & 35 $\times$ 35 $\times$ 192 \\
			3 $\times$ Inception &  Inception filters & 35 $\times$ 35 $\times$ 288 \\
			5 $\times$ Inception &  Inception filters & 17 $\times$ 17 $\times$ 768 \\
			2 $\times$ Inception &  Inception filters & 8 $\times$ 8 $\times$ 1280 \\
			pool &  8 $\times$ 8 & 8 $\times$ 8 $\times$ 2048 \\
			linear &  logits & 1 $\times$ 1 $\times$ 2048 \\
			softmax &  classifier & 1 $\times$ 1 $\times$ 1000 \\ \hline	
		\end{tabular}\hfil}
\end{table}
\subsubsection{Adversarial Attacks}
Adversarial examples are defined as malicious patterns created by the slightly modified aim to fool an AI model but indistinguishable from humans.

\textbf{FGSM (Fast Gradient Sign Method).} was proposed by Ian Goodfellow et al.~\cite{43405}. In a normal training process, the input and output data distributions are assumed as fixed and unchangeable, so there are only trainable parameters and weights that are fine-tuned respect to a loss objective function between input \(x\) and label \(y\).~\cite{43405} used a very simple idea to reverse that normal process when they fine-tuned input data distribution respect to a new loss objective function between new sample \(x^{adv}\) with new specific label \(y^{adv}\):
\begin{equation}\label{eq:fgsm}
x^{adv}= x-\beta  \cdot sign(\triangledown_xLoss(x^{adv},y^{adv}))
\end{equation}
where \(\beta \) denotes the perturbation size to create an adversarial example \(x^{adv}\) from a legitimate input \(x\). 	From a legitimate input \(x\), FGSM looks for the best adversarial perturbation \(\beta \) to add into \(x\) to create a new image \(x^{adv}\). The value of \(\beta \) has to satisfy two requirements include the magnitude of \(\beta \) is as small as possible and respect to the loss objective function between (x,y). For the first requirement, the magnitude of \( \beta \) is smaller, \(x^{adv}\) is more similar as \(x\) but the convergence rate of the algorithm~\ref{eq:fgsm} is slower, while the bigger \(\beta \) makes \(x^{adv}\) is more different from the \(x\) but the FGSM algorithm converges faster. For the second requirement, the loss objective function between (x,y) is maximized and \(Loss(x^{adv},y^{adv})\) is minimized. Because the total of probabilities of output is equal to one, so the algorithm~\ref{eq:fgsm} only needs to consider to minimize \(Loss(x^{adv},y^{adv})\). 

In this paper, we use the FGSM~\cite{43405} method with \(l\)-norm optimization as a baseline to conduct assessments of the possible value areas of \(\beta \) during the creation of adversarial examples. Our attack method is based on a white-box attack where victim AI model information is known in advance and can be accessed.

\subsubsection{Defensive approaches}
There are many methods of protection that have been proposed. The typical strategy is adversarial training~\cite{szegedy2013intriguing, kurakin2016adversarial1,tramer2017ensemble}. The idea of this strategy of protection is that the AI models will be trained with adversarial examples and ground truth labels. With the assumption that the more AI models are learned, the more accurate they will regain and the more likely it will be to misidentify adversarial examples. However, the major drawback of the adversarial training method is that it takes a lot of time to create adversarial examples and training time for AI models. In addition, this method does not guarantee resistance to new adversarial examples created by other methods than those created by the previous method. Other defensive methods that are often investigated to be pre-processing data. These defensive methods include preprocessing methods based on image transformation~\cite{DBLP:journals/corr/abs-1901-03037, 8958446}, filter~\cite{thang2019automated, dangduythang2019label} or compression~\cite{das2017keeping}. Those methods of defense have very impressive results in helping AI systems identify which input is adversarial or legitimate. 
One of the defenses which also attracts high attention is gradient masking. The adversarial attack methods are largely based on gradient calculations to optimize the loss objective function when creating adversarial examples. For that reason, the idea of hiding the gradient value was proposed.~\cite{tramer2017ensemble} proposed a gradient masking method based on smoothing the gradient gradients that made the global optimal calculation based on gradient slope is more difficult.~\cite{papernot2016distillation} uses another strategy that is distillation synthesized from different models to create a stronger model against adversarial examples.
\section{Search Space of Adversarial Perturbation} \label{proposedmethod}
\subsection{Search Space on Attacking Phase}
One of the important factors in the process of creating adversarial examples is the adversarial perturbation coefficient \(\beta \). 
However, how to find out the optimal value of \(\beta \) and its relationship to the currently most powerful defense methods in relation to image filter~\cite{thang2019automated} is unclear. That is the purpose of this study. In this research, we investigate on a white-box attack in creating adversarial examples. This is the setting defined as the attacker can access and use the AI model parameters for conducting an attack pattern. This is possible because currently, the most powerful AI models in image classification tasks are open-source. Many attack methods have been proposed, but most of them rely on FGSM for development, generally, we also use FGSM for creating adversarial examples. One thing to note, it is possible to classify adversarial attacks into two different types based on the purpose of the attacker include non-targeted and targeted attacks. The non-targeted attack is defined as the attacker only focuses on maximize the loss function of \((x,y)\) in order to deceive the AI system. Meanwhile, a targeted attack is defined as the attacker wants to trick the AI system into a misclassifying new pattern in an intentional label rather than merely misidentifying it. Because of this, targeted attacks are more commonly used than non-targeted attacks and we also use it in attacking phases.
Our main purpose to decide the size of adversarial perturbation, it means the search space of adversarial perturbation. We consider the norm operation to determine the size of the adversarial noises. Mathematically, the norm operation is used to calculate the distance, or the length of the vectors or the matrixes according to element-wise. The bigger the norm value, the bigger the difference between vectors or matrices and vice versa. Formally, the \(l_p\)-norm of vector x is defined as: \(\left \| x \right \|_p = \sqrt[p]{\sum _i \left | x_i \right |^p}\), where \(p \in \mathbb{R}\). This is a \(p^{th}\)-root of a summation of all elements to the \(p^{th}\) power is what we call a norm. The important point is even though every \(l_p\)-norm is all looked very similar to each other, their mathematical properties are very different and thus their application is completely different when we use to create the adversarial examples. In this work, we consider three common norm methods: \(l_1\)-norm, \(l_2\)-norm, and \(l_{\infty }\)-norm for evaluating the size of the search space of adversarial perturbation. 

\textbf{\(\mathbf{l_1}\)-norm.} We define \(x_{true}\) as the original input vector, \(l_1\)-norm of \(x_{true}\) is defined as:
\begin{equation}
\left \| x_{true} \right \|_1 = \sum _i \left | x_{true}^{(i)} \right |
\end{equation}
This norm is also well-known as the Manhattan norm and it is one of very common norm operations.

\textbf{\(\mathbf{l_2}\)-norm.} is the most popular norm and also known as the Euclidean norm. 
The \(l_2\)-norm and other norms are equivalent in the sense that all of them are defined in the same topology. The \(l_2\)-norm is defined as:
\begin{equation}
\left \| x_{true} \right \|_2 = \sqrt{\sum _i(x_{true}^{(i)})^2}
\end{equation}
We use the \(l_2\)-norm to measure the difference between two vectors \(x_{true}\) and \(x_{adv}\), the \(l_2\)-norm is re-defined:
\begin{equation}
\left \| x_{true} - x_{adv} \right \|_2 = \sqrt{\sum _i(x_{true}^{(i)} - x_{adv}^{(i)})^2}
\end{equation}
where \(x_{adv}\) defines the adversarial example.

\textbf{\(\mathbf{l_{\infty}}\)-norm.} The \(l_{\infty}\)-norm is defined as equation below:
\begin{equation}
\left \| x_{true} \right \|_{\infty } = \sqrt[\infty ]{\sum_{i}(x_{true}^{(i)})^{\infty}}
\end{equation}
Let consider the vector \(x\), if \(x^{(i)}\) is each element in vector \(x\), from the property of the infinity itself, we have: \(x_i^{\infty } \approx x_k^{\infty }\forall i\neq k\), then \(\sum_{i}x_i^{\infty }=x_k^{\infty }\). And we have \(\left \| x \right \|_\infty =\sqrt[\infty ]{\sum_{i}x_i^\infty}=\sqrt[\infty ]{x_k^\infty}=\left | x_k \right |\). Now we have simple definition of \(l_{\infty}\)-norm as: \(\left \| x \right \|_\infty =max(\left | x_i \right |)\). 

So our attack phase is denoted as Algorithm~\ref{alg:attack_phase} by using FGSM. Where \(x_{true}\) defines the original input, \(x_{adv}\) is adversarial example, \(y_{true}\) defines the ground-truth label, \(y_{adv}\) is an adversarial label, f is the activation function of machine learning model, \(\beta \) is the maximum adversarial value, \(l_i\) defines the norm. For crafting adversarial example, we set a learning rate \(lr\) is equal to 0.01, the number of iteration is 500 times. 
\begin{algorithm}[h]
	\SetKwData{Left}{left}
	\SetKwData{This}{this}
	\SetKwData{Up}{up}
	\SetKwFunction{Union}{Union}
	\SetKwFunction{FindCompress}{FindCompress}
	\SetKwInOut{Input}{input}
	\SetKwInOut{Output}{output}
	\Input{$x_{true}$, $y_{true}$, $y_{adv}$, $f$, $\beta$, $l_i$}
	\Output{$x_{adv}$}
	\Parameter{lr = 0.01, iterations = 500}
	\BlankLine
	$x_{adv}\leftarrow x_{true}$	\tcp{initial adversarial example}
	$\delta \leftarrow \vec{0}$ \tcp{initial adversarial perturbation}
	$it \leftarrow 1$ \tcp{initial iteration loop}
	\While{$\delta$ $< \beta$ and $f(x_{adv})\neq y_{adv}$ and $it <= iterations$}{
		$x_{adv} \leftarrow x_{true}-\delta \cdot sign(\bigtriangledown Loss(y_{adv}|x_{adv}))$\\
		$\delta \leftarrow $ norm($l_i$)\\
		maximize $Loss(y_{adv}|x_{adv})$ respect to $\delta$\\
		$\delta \leftarrow clip(x_{adv},x-\beta,x+\beta)$\\
		$it \leftarrow it + 1$ \\
	}
	\KwRet{$x_{adv}$}
	\caption{Crafting Adversarial Examples with \(l\)-norm optimization}\label{alg:attack_phase}
\end{algorithm}
\subsection{Filter Methods}
Most adversarial attack methods look for the optimal values of adversarial perturbation respect to loss objective function to modify the original image. Therefore, the pixels that are incidentally edited are located in the high-frequency domain. Therefore current protection methods based on image filters have proved very effective in eliminating these adversarial noises. However, in order to better understand the search space of this adversarial perturbation and the ability to resist image filters, we studied the two most common image filters, the Gaussian and the Median filter. Mathematically, a Gaussian filter modifies the input image by calculating a convolution the area of a specific image area with a Gaussian function; this transformation is also known as the Weierstrass transform. The area of convolution is often called kernel size and is usually 3x3 or 5x5. When using a Gaussian filter, the kernel window will move across the surface of the input image and compute the kernel window that corresponds to the image area being processed. The second image filter to be considered in this research is the median filter. This is a very common filter used to highlight the edges of an image. The Median filter also uses kernel windows that move across the input image surface. However, the median filter processes that area simply by finding the median value of the image area being processed, then replacing that median value in the pixel position in the center of the windows kernel while preserving the pixel values in neighbors.
Our filtering system proceeds by Algorithm~\ref{alg:detection_system}, where x defines the input image, \(\varphi\) denotes the kernel sizes, \(f\) is a machine learning function that computes the predicted label with the highest probability, \(y_{true}\) defines the ground truth label, \(y_{adv}\) defines the adversarial label, and \(s\) is the filter function. Output are the probabilities of the ground truth label (\(p_{true}\)) and the adversarial label (\(p_{adv}\)).
\begin{algorithm}[h]
	\SetKwData{Left}{left}
	\SetKwData{This}{this}
	\SetKwData{Up}{up}
	\SetKwFunction{Union}{Union}
	\SetKwFunction{FindCompress}{FindCompress}
	\SetKwInOut{Input}{input}
	\SetKwInOut{Output}{output}
	\Input{$x_{true}, s, f, y_{true}, y_{adv}$} 
	\Output{$p_{true}, p_{adv}$}
	\Parameter{$\varphi =[(3\times3);(5 \times 5)]$}
	\BlankLine
	\For{$i$ in $\varphi $}{
		$x_{filtered}$ $\leftarrow$ $s(x, i)$\\
		$P$ $\leftarrow$ $f(x_{filtered})$\\
		$p_{true}$ $\leftarrow$ $P(y_{true})$\\
		$p_{adv}$ $\leftarrow$ $P(y_{adv})$\\
	}
	\caption{Image Filters on input for image classification task}\label{alg:detection_system}
\end{algorithm}
\section{Implementation and Results}\label{experiements}
\subsection{Datasets and AI model}
The target AI model that we use in the implementation is Google Inception V3~\cite{szegedy2016rethinking} that was trained with 1,000 common categories in the ImageNet~\cite{russakovsky2015imagenet} dataset. Our attacking phase is a white-box attack setting and a targeted label is ``street sign'' label. We use FGSM with \(l_1\)-norm, \(l_2\)-norm, and \(l_\infty \)-norm to craft adversarial images.
\subsection{Results}
Intuitively, because of the copyright issue of the ImageNet dataset, we use our own images (include pictures of vending machine, computer mouse and keyboard) for analysis. We randomly selected targeted labels for the creation of adversarial images. By using the FGSM method in combination with \(l_1\)-norm, \(l_2\)-norm, and \(l_\infty \)-norm, from each original image we create three different adversarial images. 

\begin{figure}[h]
	\centering
	\includegraphics[width=0.45\textwidth]{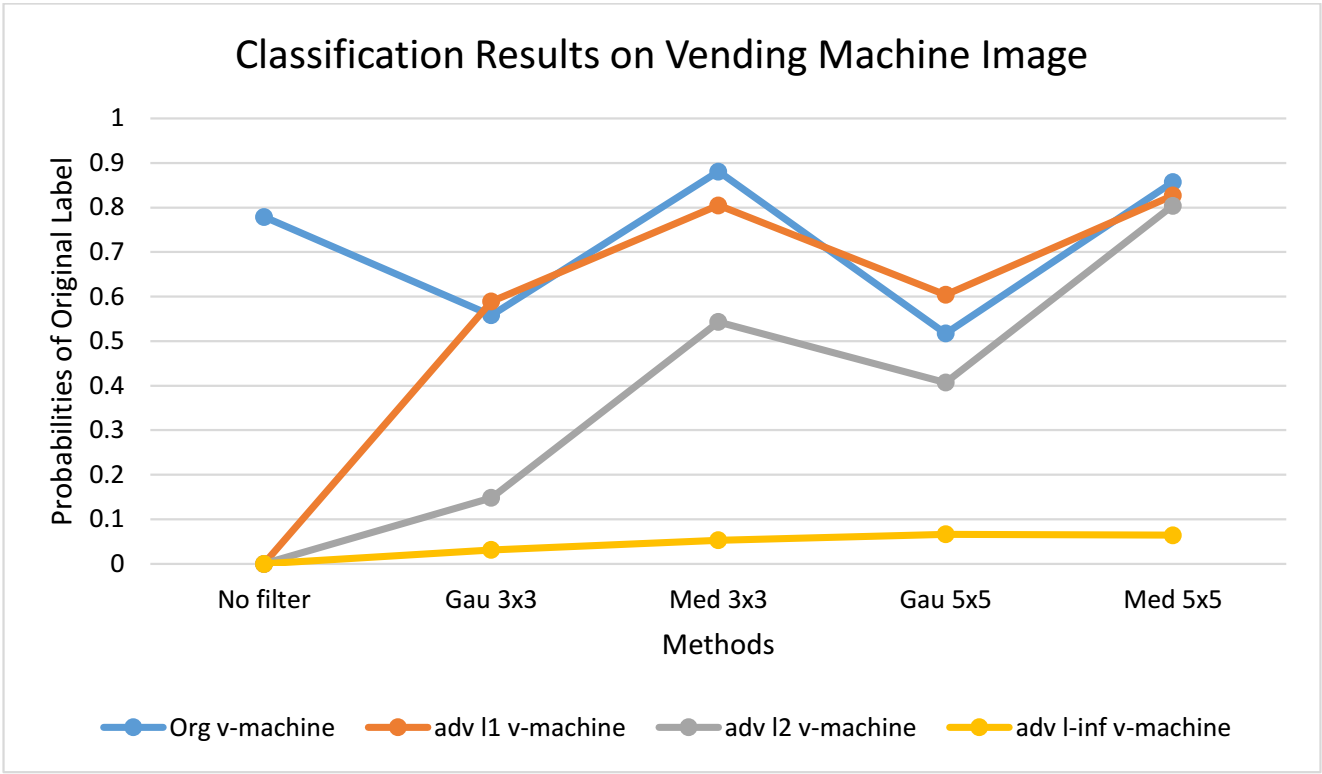}
	\caption{Classification Results on Vending Machine Image with observation on the probabilities of Original Label}
	\label{fig:eval_org_label_v_machine}
\end{figure}

\begin{figure}[h]
	\centering
	\includegraphics[width=0.45\textwidth]{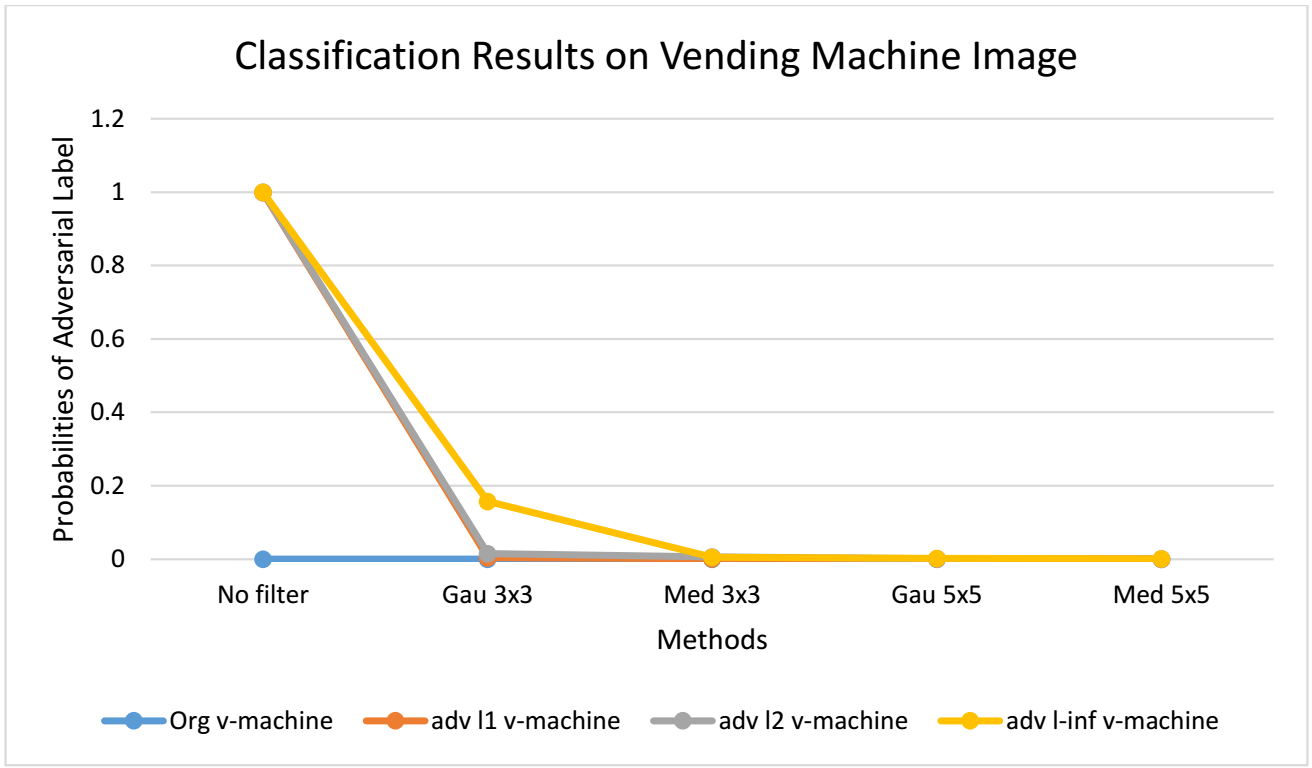}
	\caption{Classification Results on Vending Machine Image with observation on the probabilities of Adversarial Label}
	\label{fig:eval_adv_label_v_machine}
\end{figure}

\begin{figure*}[h]%
	\centering
	
	\subfloat[Observation on the probabilities of Original Keyboard Label]{{\includegraphics[width=0.45\textwidth]{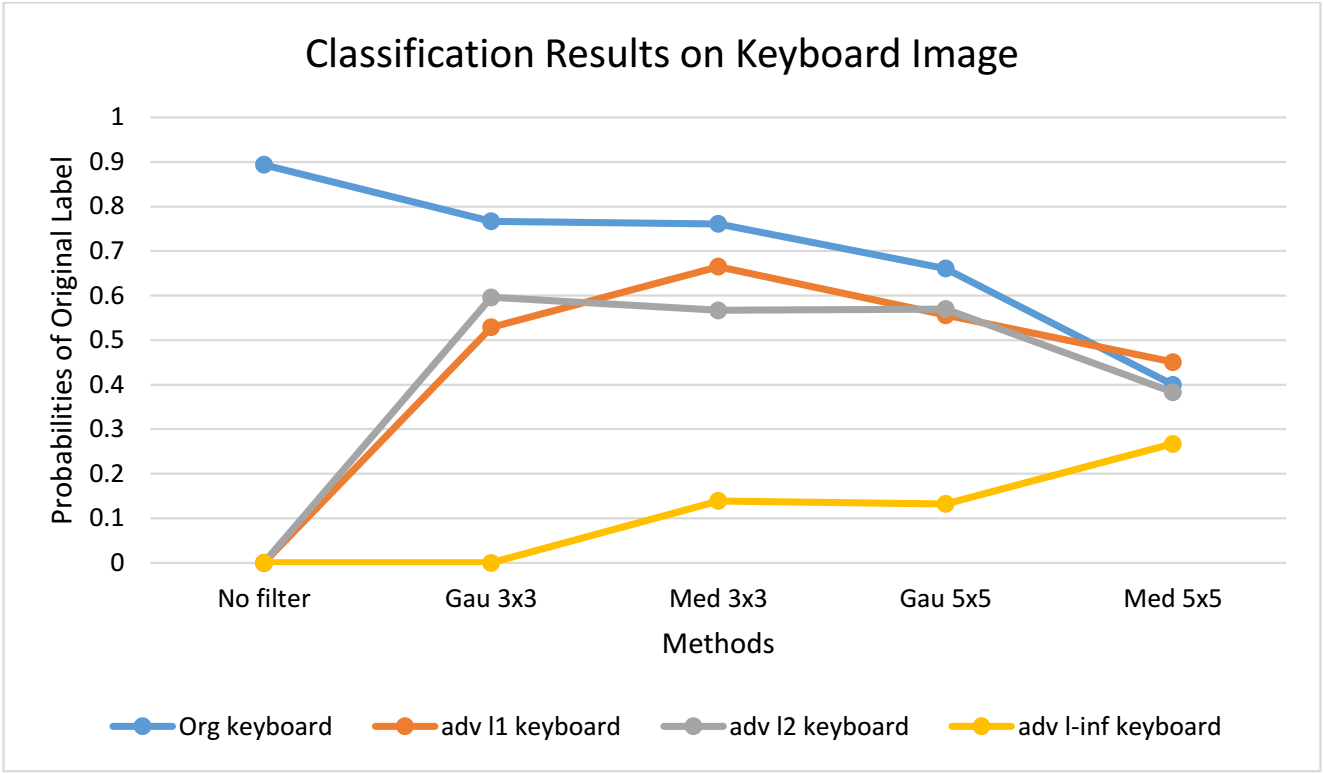} }}%
	\qquad
	\subfloat[Observation on the probabilities of Adversarial Label]{{\includegraphics[width=0.45\textwidth]{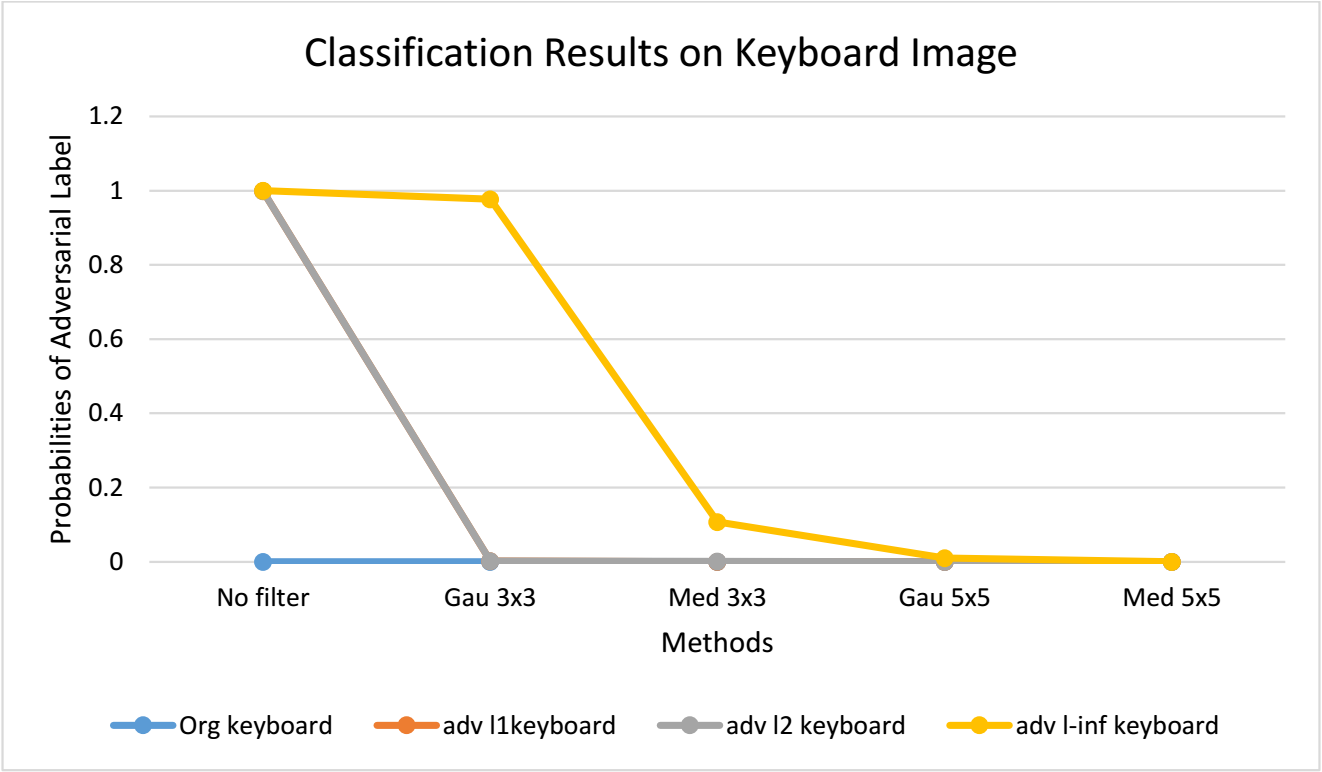} }}%
	\qquad
	\subfloat[Observation on the probabilities of Original Computer Mouse Label]{{\includegraphics[width=0.45\textwidth]{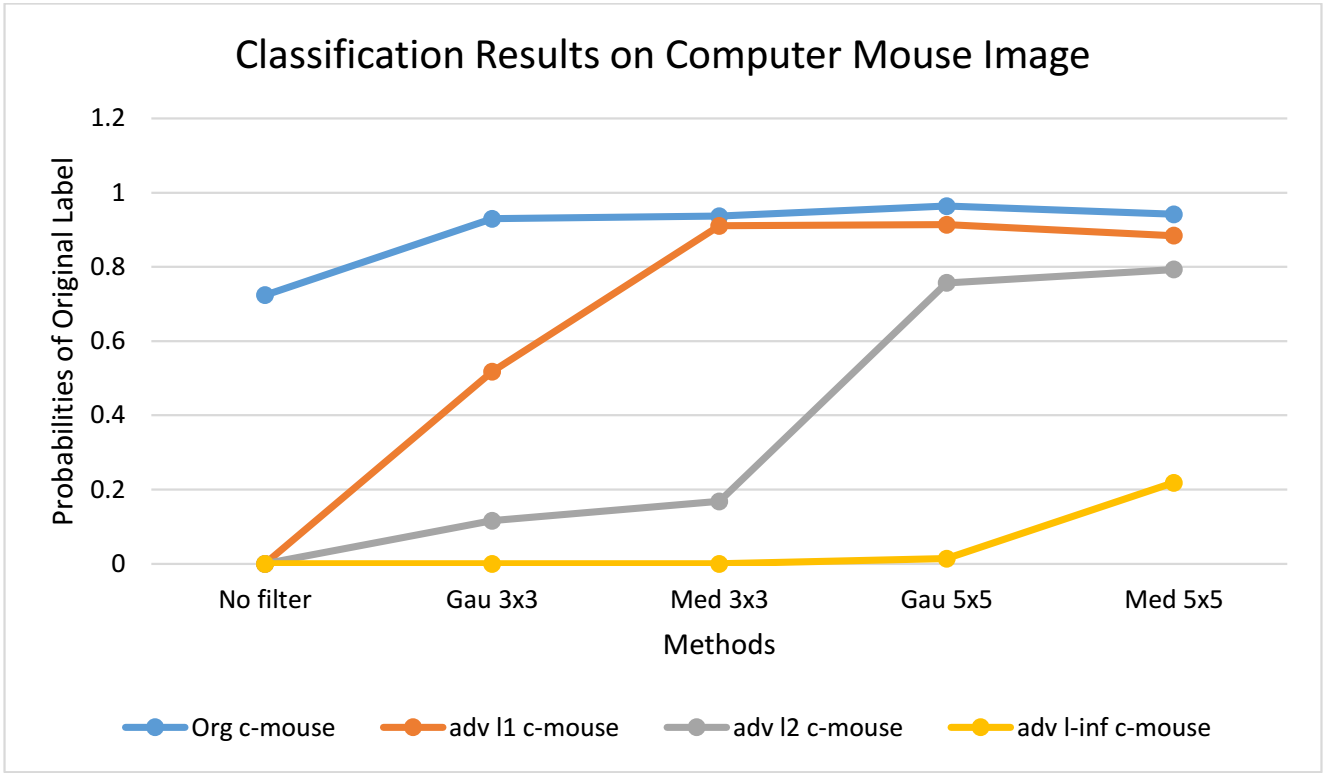} }}%
	\qquad
	\subfloat[Observation on the probabilities of Adversarial Label]{{\includegraphics[width=0.45\textwidth]{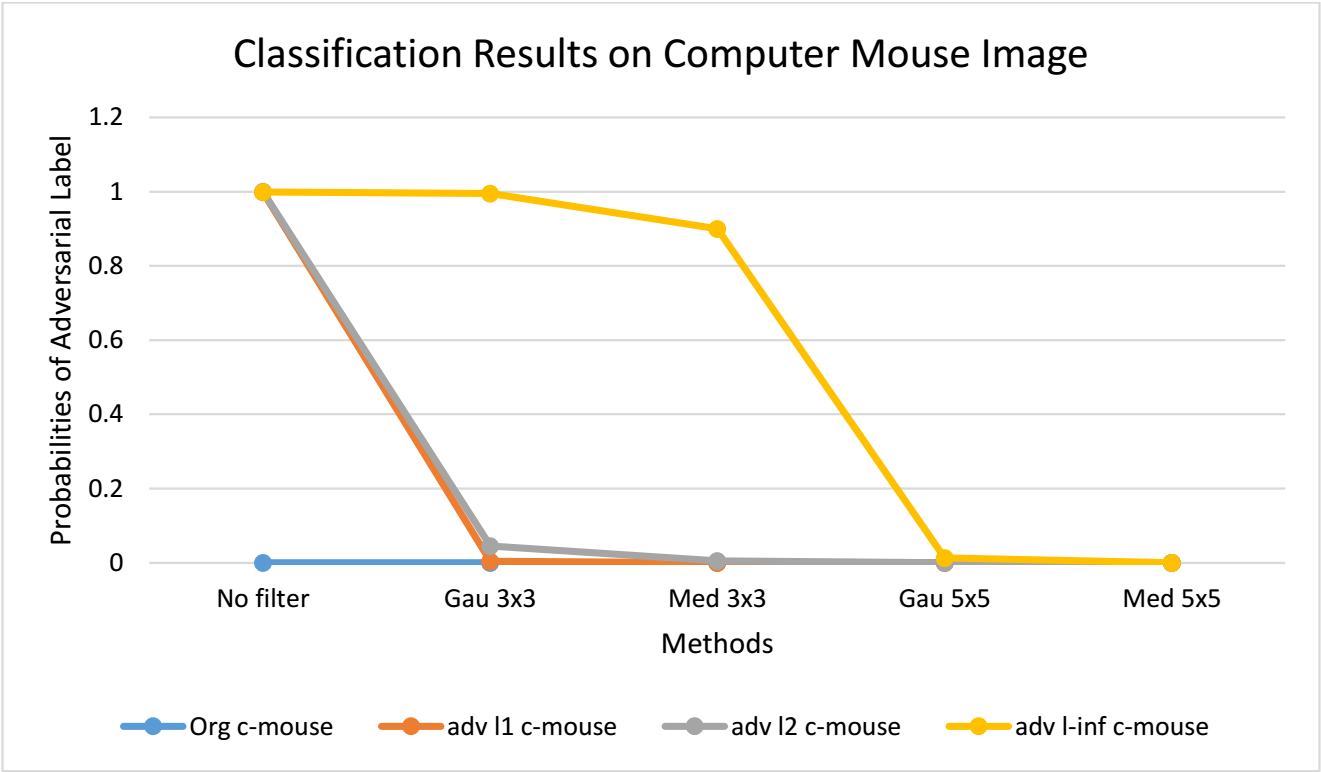} }}%
	\caption{Classification Results on Keyboard and Computer Mouse Images}%
	\label{fig:eval_keyboard_mouse}%
\end{figure*}

Fig.~\ref{fig:eval_org_label_v_machine} shows the probabilities of the original vending machine label when the input is an vending machine image. 
Fig.~\ref{fig:eval_adv_label_v_machine} shows the probabilities of the adversarial label with vending machine input.  
Fig.~\ref{fig:eval_keyboard_mouse} shows the observations on the images of computer mouse and keyboard.

\begin{figure*}[h]%
	\centering

	\subfloat[Orginal Vending Machine]{{\includegraphics[width=0.45\textwidth]{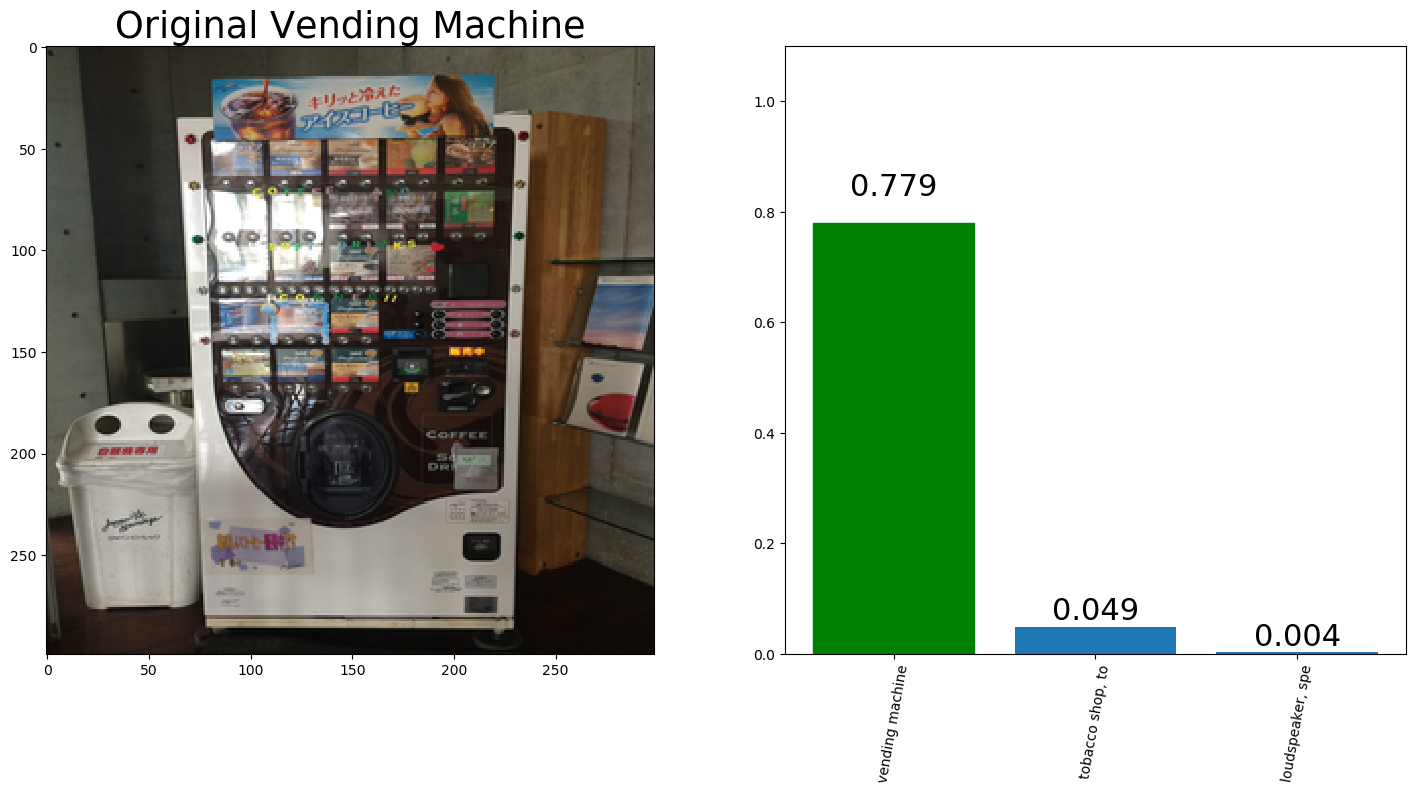} }}%
	\qquad
	\subfloat[Adversarial FGSM\_L\(_1\) Vending Machine]{{\includegraphics[width=0.45\textwidth]{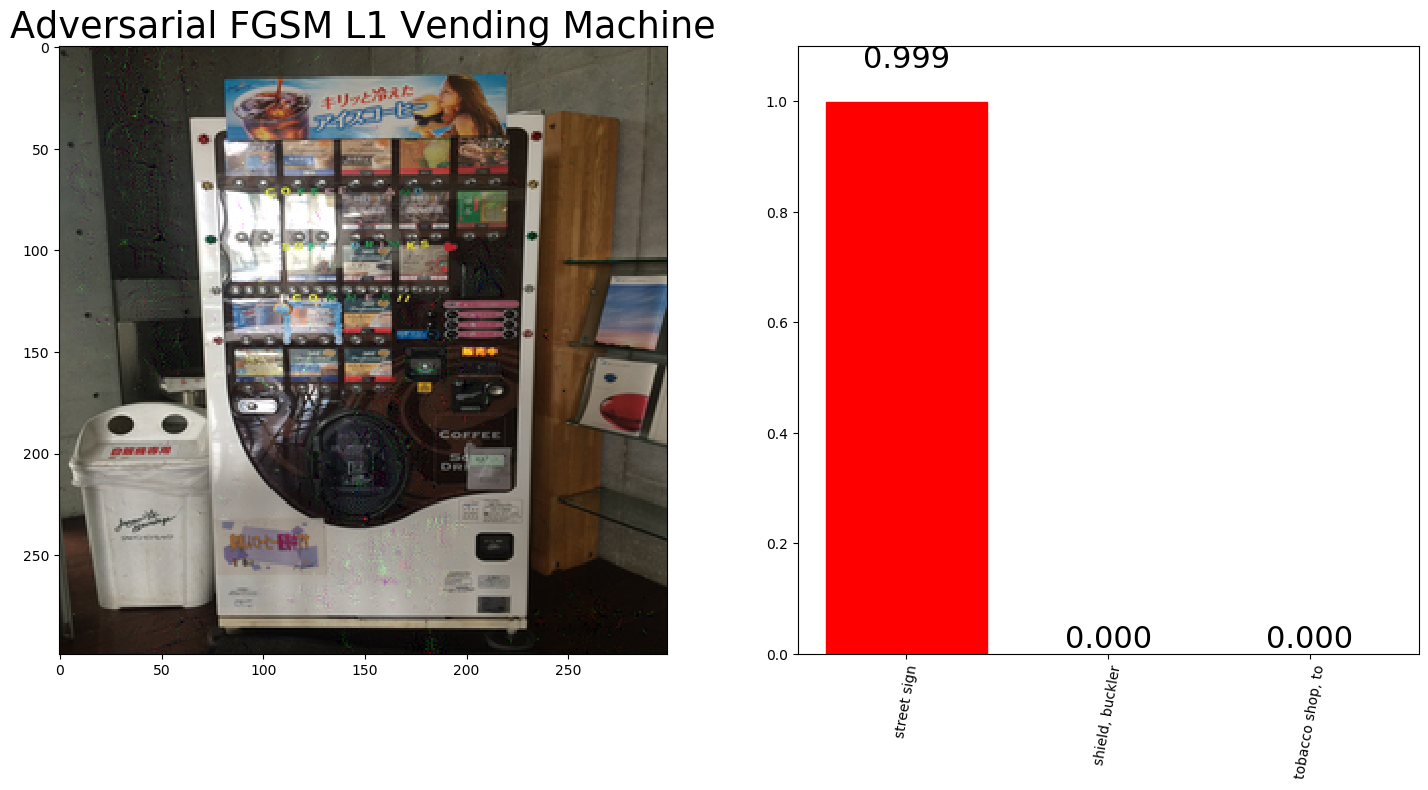} }}%
	\qquad
	\subfloat[Adversarial FGSM\_L\(_2\) Vending Machine]{{\includegraphics[width=0.45\textwidth]{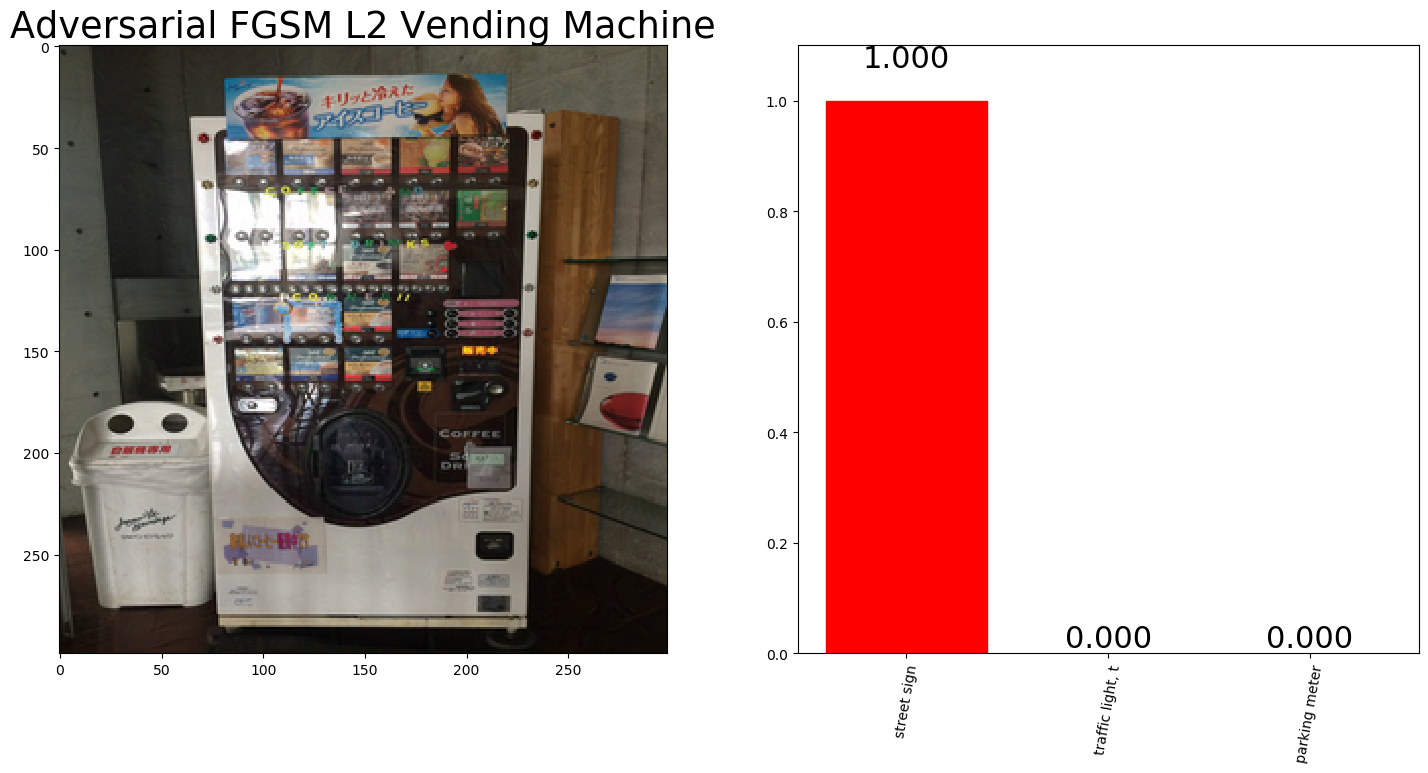} }}%
	\qquad
	\subfloat[Adversarial FGSM\_L\(_{\infty}\) Vending Machine]{{\includegraphics[width=0.45\textwidth]{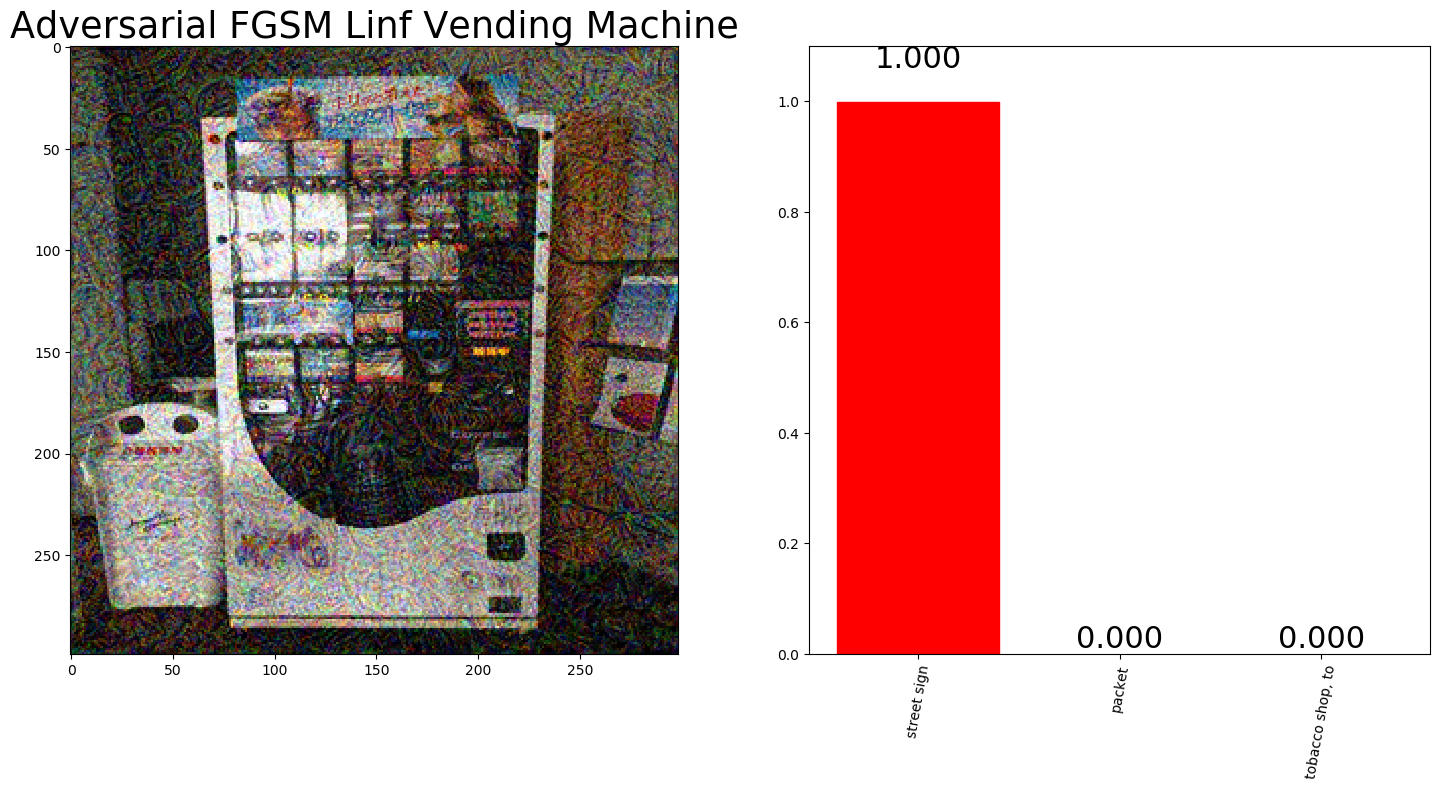} }}%
	\caption{Adversarial Vending Machine (targeted class: Street Sign)}%
	\label{fig:vending_machine_adv}%
\end{figure*}

Fig.~\ref{fig:vending_machine_adv} shows the results of creating adversarial images from the original image of the vending machine. We find that the deep learning system is easily fooled with adversarial images. In addition, we intuitively observe that adversarial images created with \(l_1\)-norm and \(l_2\)-norm are harder to detect than \(l_\infty \)-norm.

\begin{figure*}[h]%
	\centering
	\captionsetup{justification=centering}
	\subfloat[Orginal Vending Machine with Gaussian \((3\times3)\) ]{{\includegraphics[width=0.45\textwidth]{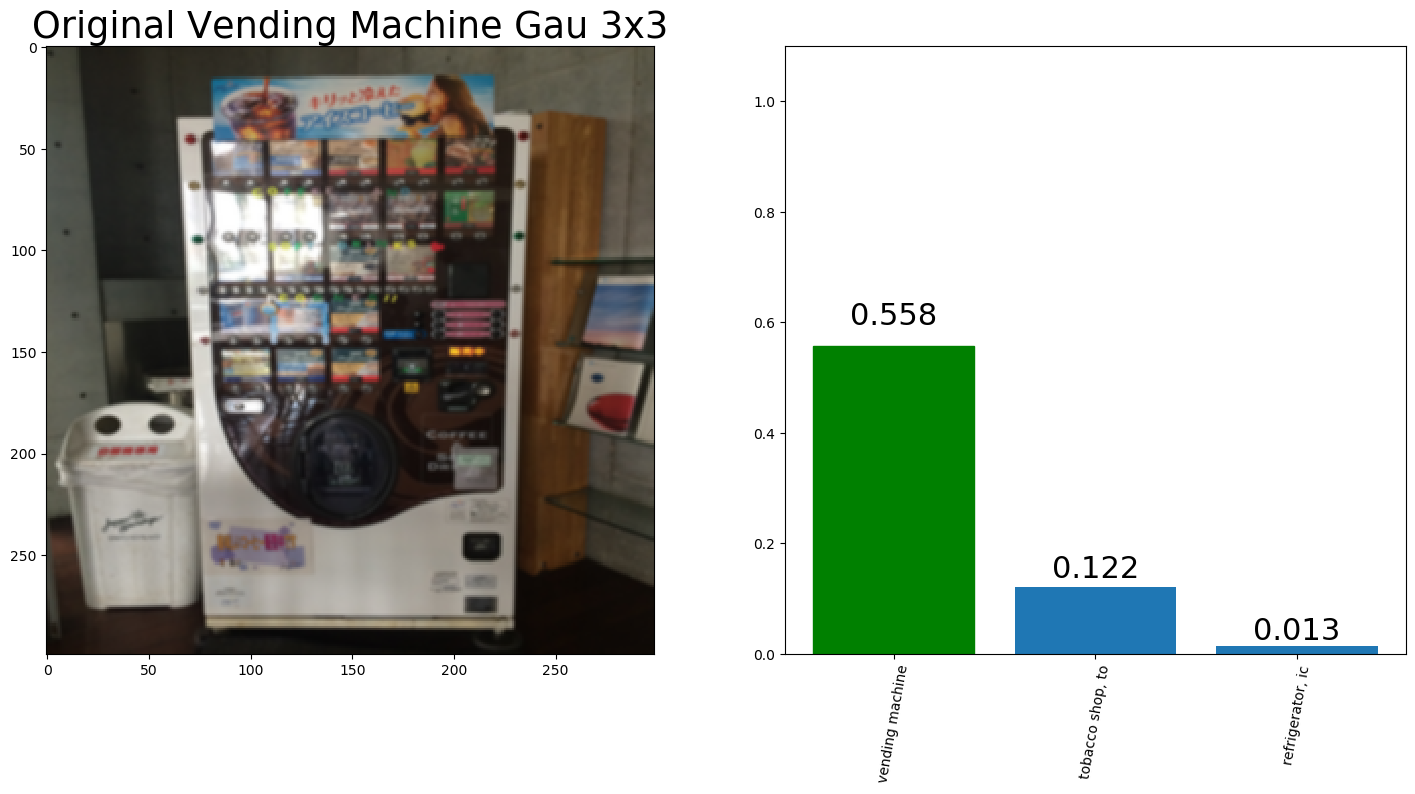} }}%
	\qquad
	\subfloat[Orginal Vending Machine with Gaussian \((5\times5)\)]{{\includegraphics[width=0.45\textwidth]{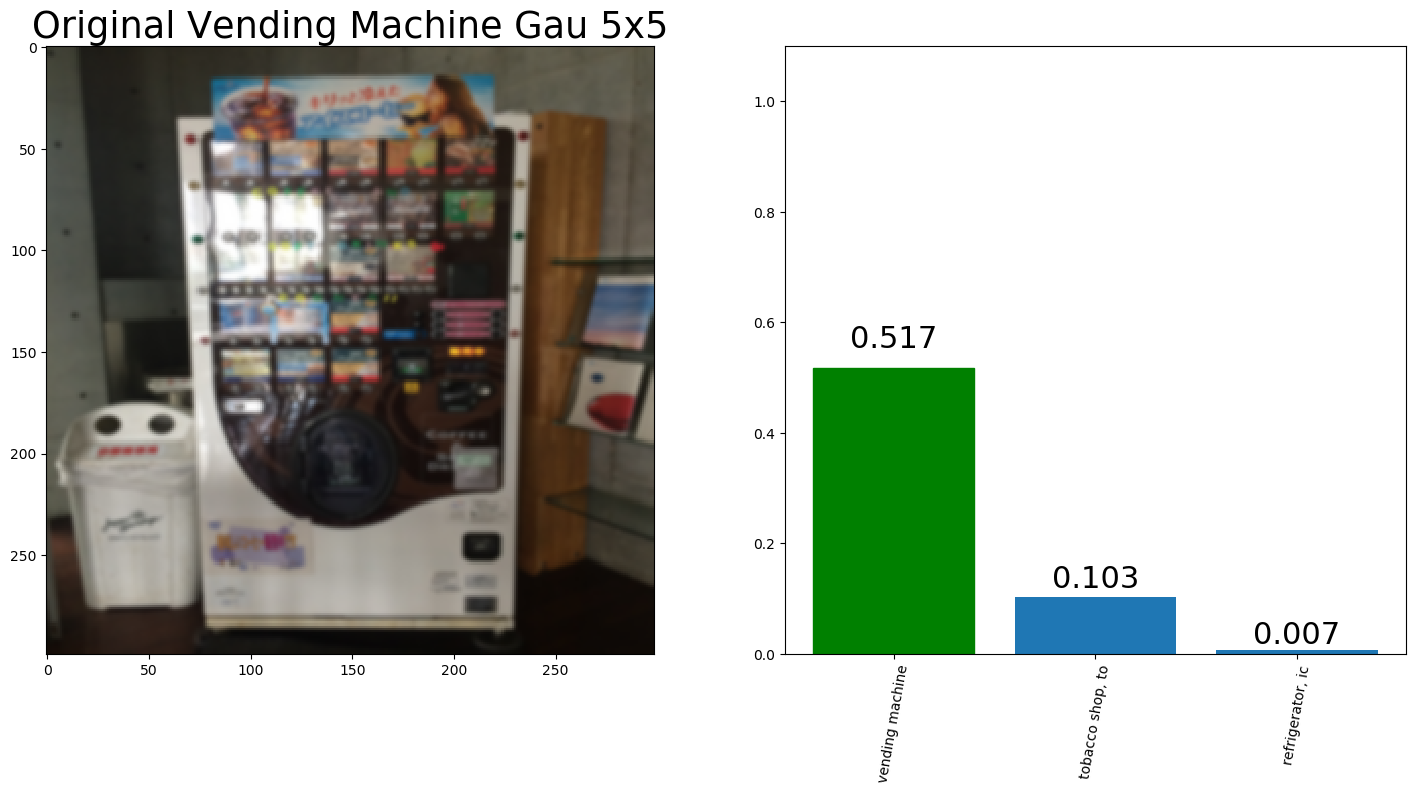} }}%
	\qquad
	\subfloat[Orginal Vending Machine with Median \((3\times3)\)]{{\includegraphics[width=0.45\textwidth]{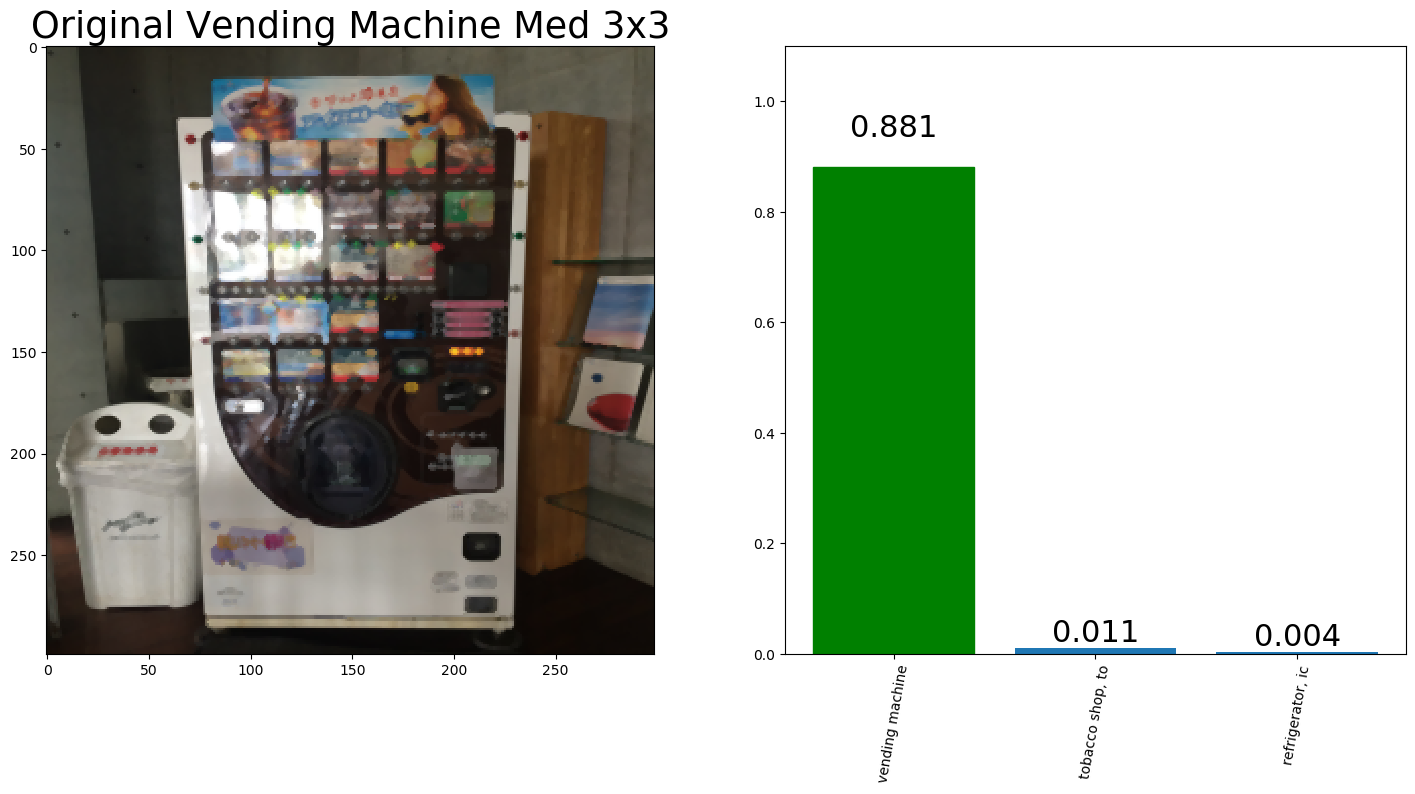} }}%
	\qquad
	\subfloat[Orginal Vending Machine with Median \((5\times5)\)]{{\includegraphics[width=0.45\textwidth]{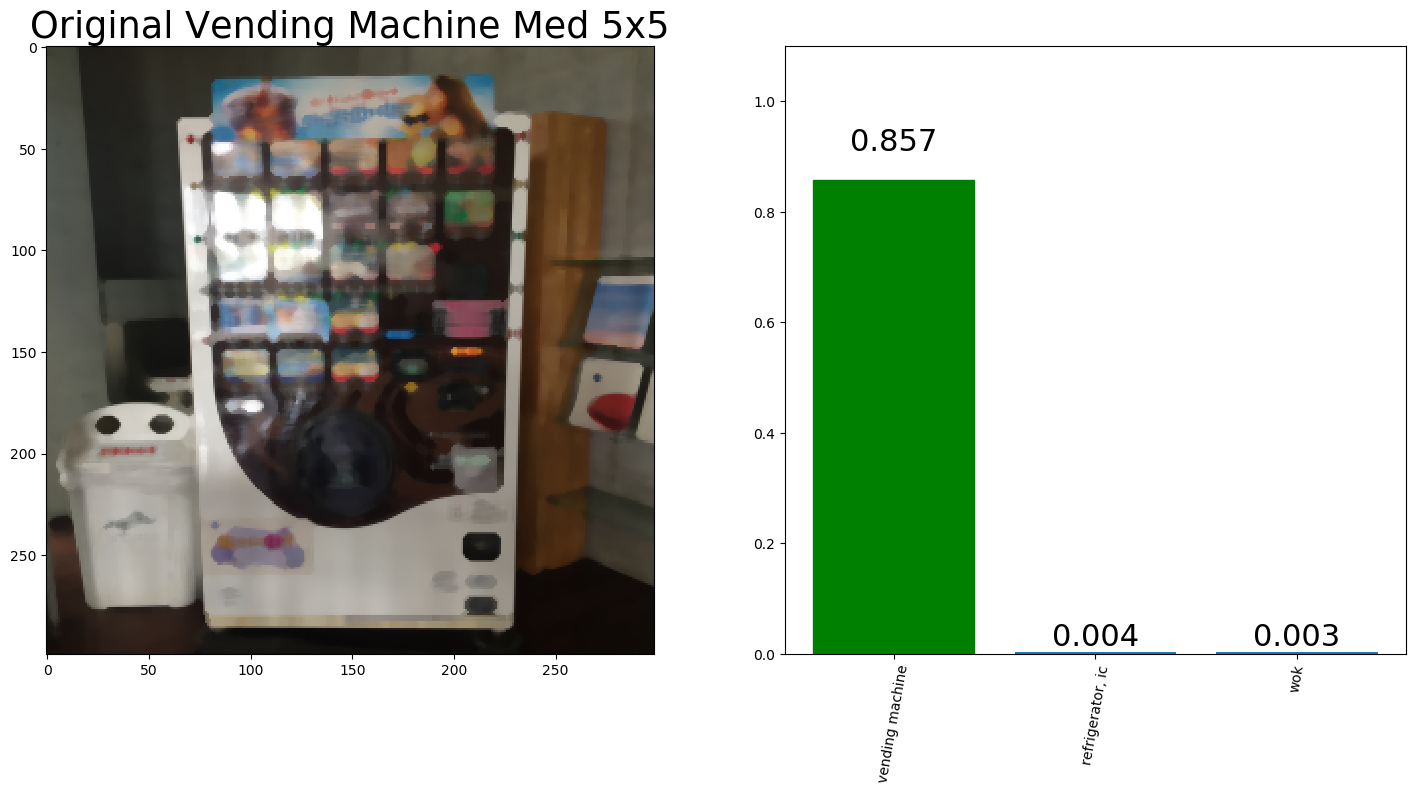} }}%
	\caption{Original Vending Machine with Image Filters}%
	\label{fig:org_vending_machine_filters}%
\end{figure*}

Fig.~\ref{fig:org_vending_machine_filters} shows the experimental results when we use the image filters method on the original image of the vending machine. We find that the Gaussian filter reduces classification accuracy more than the median filter. Especially in the case of the median with size filter 3\(\times \)3 and 5\(\times \)5, the classification results are better than the original image.

\begin{figure*}[h]%
	\centering
	\captionsetup{justification=centering}
	\subfloat[Adversarial Vending Machine with Gaussian \((3\times3)\) ]{{\includegraphics[width=0.45\textwidth]{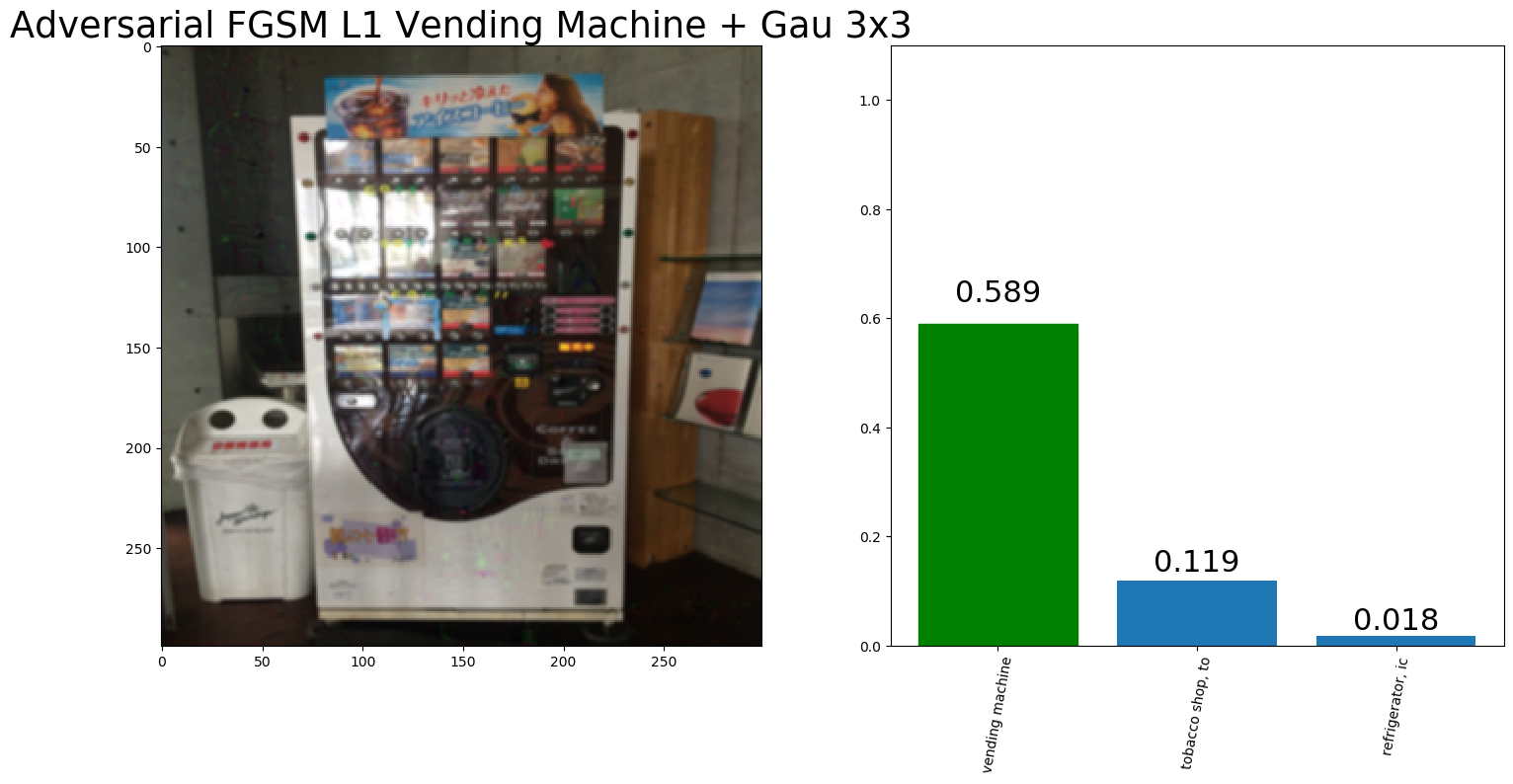} }}%
	\qquad
	\subfloat[Adversarial Vending Machine with Gaussian \((5\times5)\)]{{\includegraphics[width=0.45\textwidth]{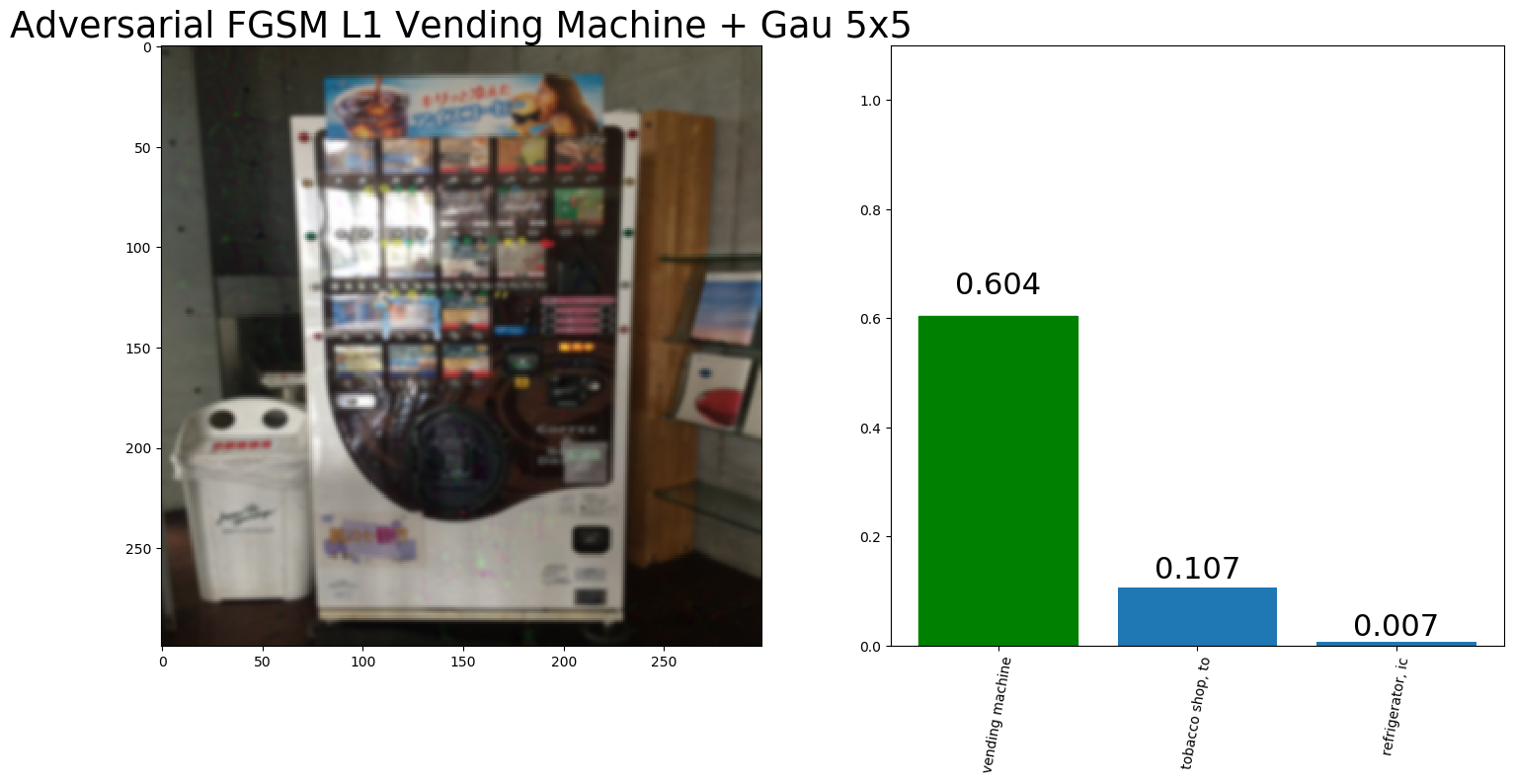} }}%
	\qquad
	\subfloat[Adversarial Vending Machine with Median \((3\times3)\)]{{\includegraphics[width=0.45\textwidth]{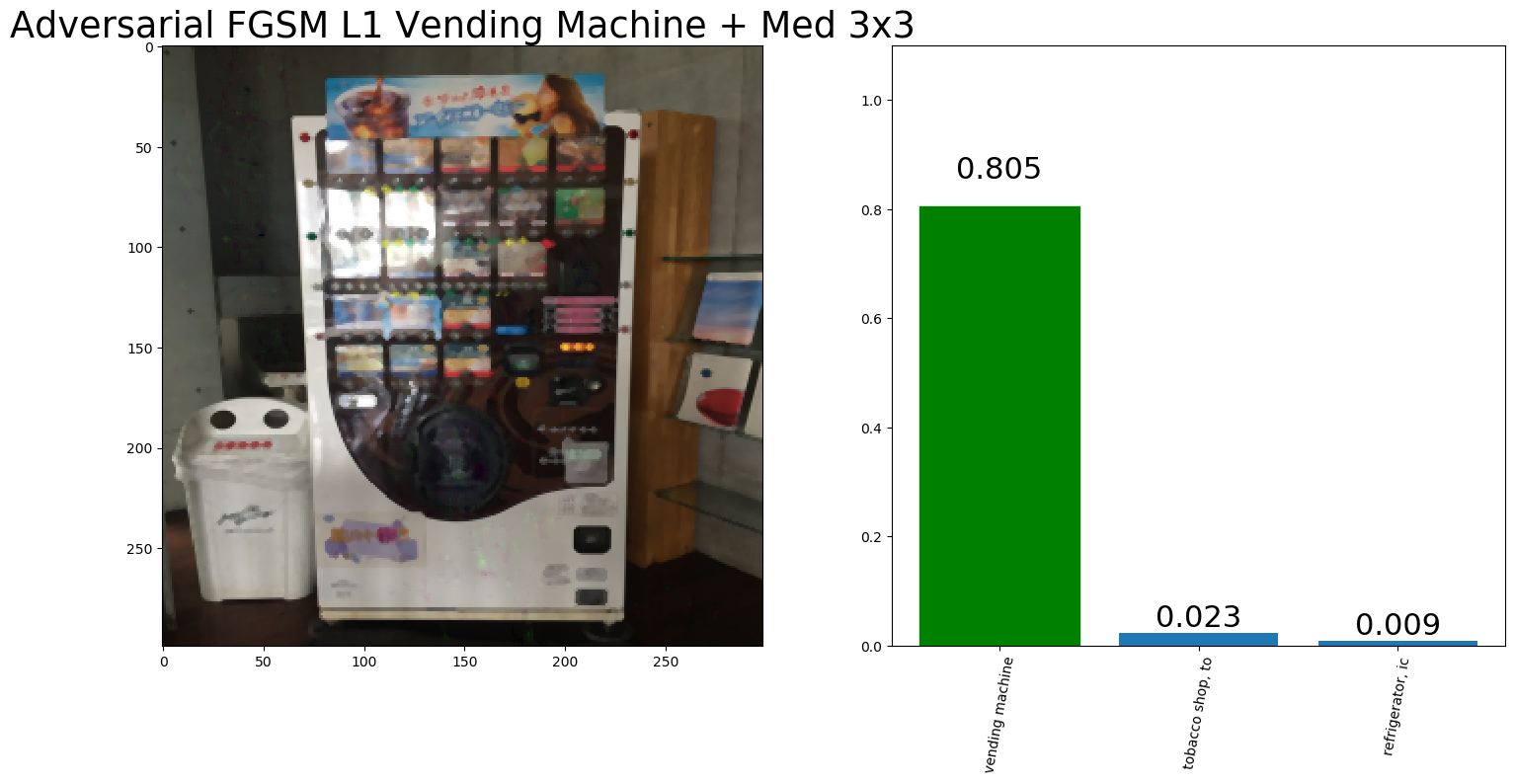} }}%
	\qquad
	\subfloat[Adversarial Vending Machine with Median \((5\times5)\)]{{\includegraphics[width=0.45\textwidth]{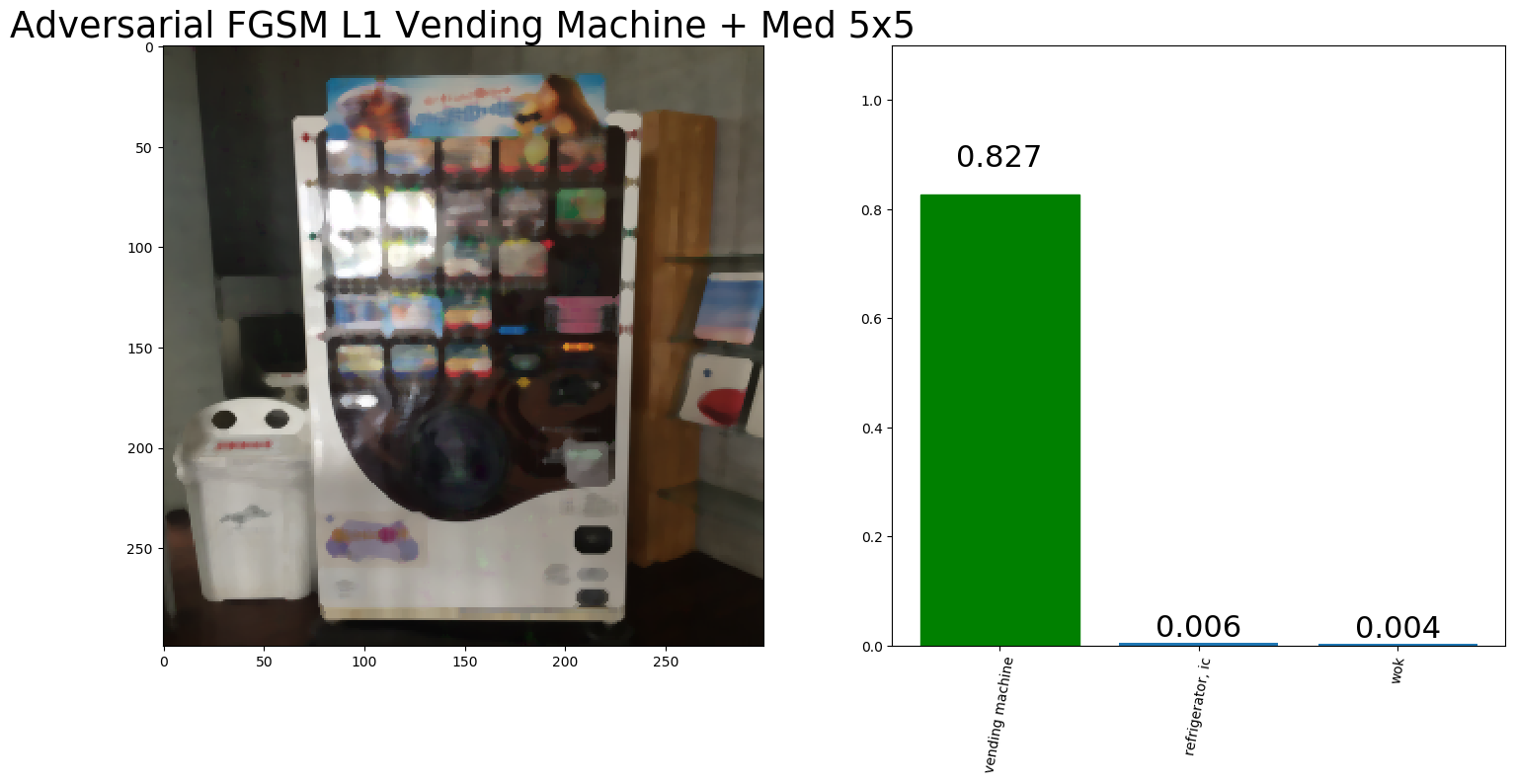} }}%
	\caption{Adversarial FGSM L\(_1\) Vending Machine (targeted class: Street Sign) with Image Filters}%
	\label{fig:adv_l1_vending_machine_filters}%
\end{figure*}

\begin{figure*}[h]%
	\centering
	\captionsetup{justification=centering}
	\subfloat[Adversarial Vending Machine with Gaussian \((3\times3)\) ]{{\includegraphics[width=0.45\textwidth]{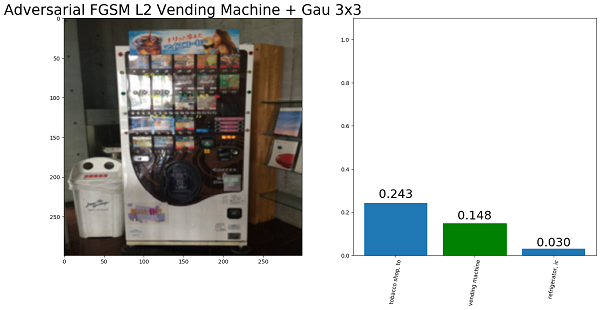} }}%
	\qquad
	\subfloat[Adversarial Vending Machine with Gaussian \((5\times5)\)]{{\includegraphics[width=0.45\textwidth]{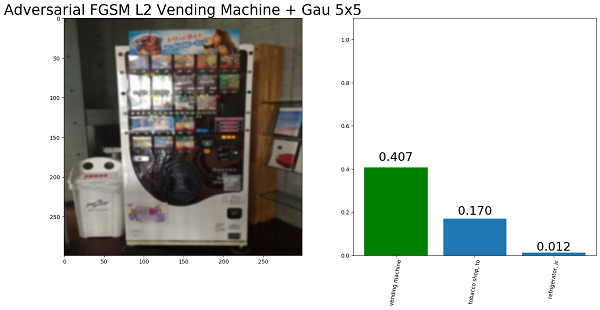} }}%
	\qquad
	\subfloat[Adversarial Vending Machine with Median \((3\times3)\)]{{\includegraphics[width=0.45\textwidth]{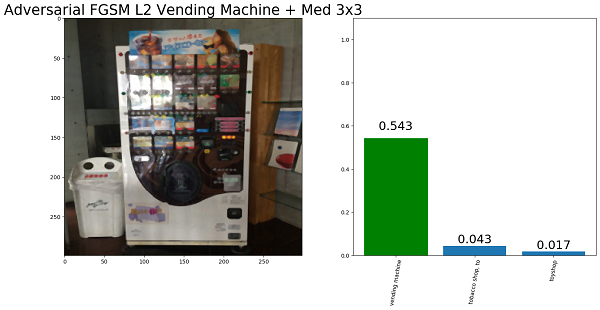} }}%
	\qquad
	\subfloat[Adversarial Vending Machine with Median \((5\times5)\)]{{\includegraphics[width=0.45\textwidth]{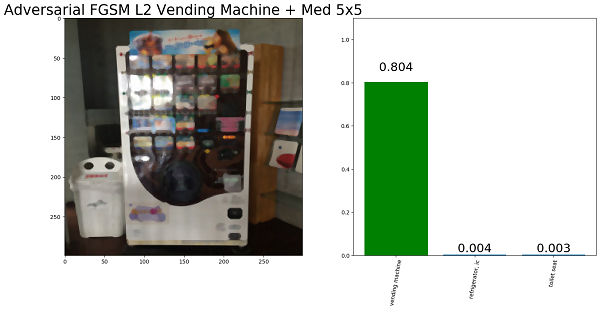} }}%
	\caption{Adversarial FGSM L\(_2\) Vending Machine (targeted class: Street Sign) with Image Filters}%
	\label{fig:adv_l2_vending_machine_filters}%
\end{figure*}

\begin{figure*}[h]%
	\centering
	\captionsetup{justification=centering}
	\subfloat[Adversarial Vending Machine with Gaussian \((3\times3)\) ]{{\includegraphics[width=0.45\textwidth]{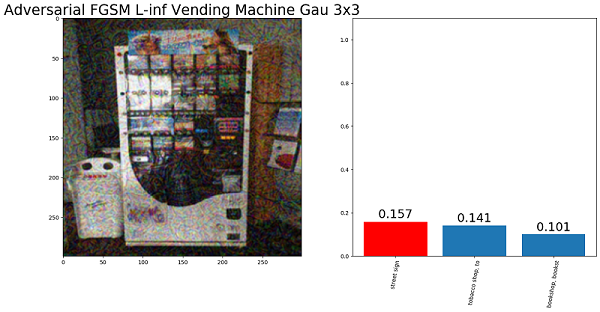} }}%
	\qquad
	\subfloat[Adversarial Vending Machine with Gaussian \((5\times5)\)]{{\includegraphics[width=0.45\textwidth]{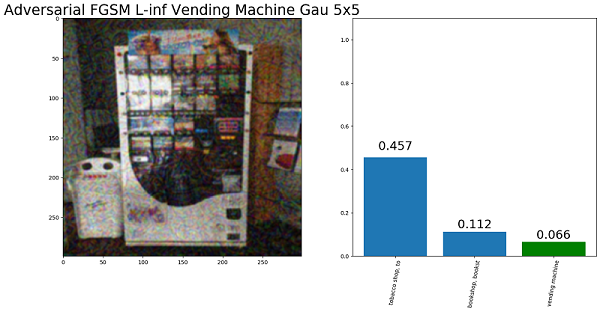} }}%
	\qquad
	\subfloat[Adversarial Vending Machine with Median \((3\times3)\)]{{\includegraphics[width=0.45\textwidth]{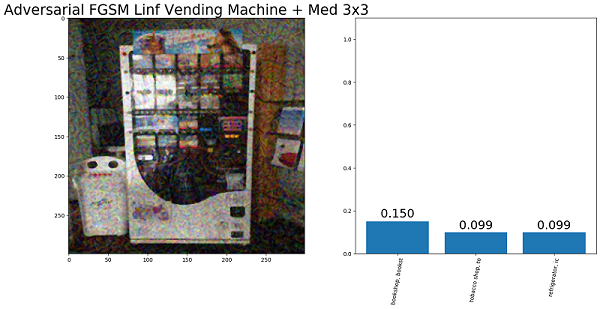} }}%
	\qquad
	\subfloat[Adversarial Vending Machine with Median \((5\times5)\)]{{\includegraphics[width=0.45\textwidth]{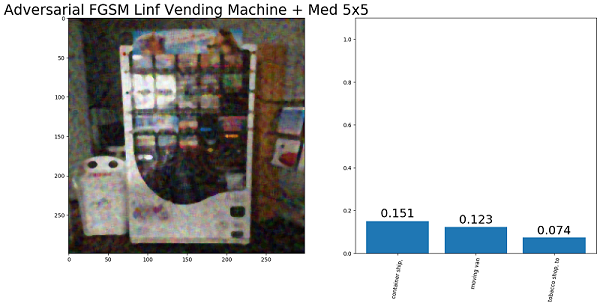} }}%
	\caption{Adversarial FGSM L\(_\infty \) Vending Machine (targeted class: Street Sign) with Image Filters}%
	\label{fig:adv_linf_vending_machine_filters}%
\end{figure*}

Similar to the original image, we also apply image filter methods to adversarial images. Fig.~\ref{fig:adv_l1_vending_machine_filters} shows classification results on adversarial images created by the FGSM method in combination with \(l_1\)-norm.
Fig.~\ref{fig:adv_l2_vending_machine_filters} illustrates classification results on adversarial images created by the FGSM method in combination with \(l_2\)-norm. We observed that Gaussian kernel size 3x3 could not restore identity to ground truth label on adversarial image with \(l_2\)-norm. The probability for vending machine label is only 14.8\%. Meanwhile, the median filter still works effectively in removing adversarial noises.
Fig.~\ref{fig:adv_linf_vending_machine_filters} shows classification results on adversarial images created by the FGSM method in combination with \(l_\infty\)-norm. We observed that Gaussian kernel size 3\(\times\)3 could not eliminate the effect of adversarial noise with \(l_\infty\)-norm on deep learning system classification. Gaussian 5\(\times\)5 gives better results, but the label with the highest probability of identification is ``tabacco shop''. The Median filter removes adversarial noises but cannot help the deep learning system correctly identify ground truth labels.

Table~\ref{tab:results} shows experimental results on vending machine (v-machine), computer mouse (c-mouse) and keyboard sets. This result shows us a large correlation between norm operations in search space of adversarial examples. It is clear that for the \(l_\infty \)-norm, the Gau (3\(\times \)3, 5\(\times \)5) and median (3\(\times \)3) methods are more difficult to completely eliminate adversarial noises based on the \(l_1\) and \(l_2\) norm. Median (5\(\times \)5) still proved superior in removing adversarial noises in all settings.
\begin{table*}[h]
	\centering
	\caption{Implementation Results}
	\label{tab:results}
	\begin{tabular}{l|l|l|l|l|l|l|l|l|l|l}
		\hline \hline
		\multirow{2}{*}{Input} & \multicolumn{2}{l|}{No filter} & \multicolumn{2}{l|}{Gau 3x3} & \multicolumn{2}{l|}{Med 3x3} & \multicolumn{2}{l|}{Gau 5x5} & \multicolumn{2}{l}{Med 5x5} \\ \cline{2-11} 
		& OL             & AL            & OL            & AL           & OL            & AL           & OL            & AL           & OL            & AL           \\ \hline \hline
		Org v-machine          & 0.779          & 0             & 0.558         & 0            & 0.881         & 0            & 0.517         & 0            & 0.857         & 0            \\ \hline
		adv \(l_1\) v-machine       & 0              & 0.999         & 0.589         & 0.004        & 0.805         & 0.001        & 0.604         & 0.001        & 0.827         & 0            \\ \hline
		adv \(l_2\) v-machine       & 0              & 1             & 0.148         & 0.015        & 0.543         & \textbf{0.006}        & 0.407         & 0.001        & 0.804         & 0            \\ \hline
		adv \(l_\infty \) v-machine     & 0              & 1             & 0.031         & \textbf{0.157}        & 0.053         & 0.005        & 0.066         & \textbf{0.002}        & 0.064         & \textbf{0.001}        \\ \hline \hline
		Org keyboard           & 0.894          & 0             & 0.767         & 0            & 0.761         & 0            & 0.661         & 0            & 0.4           & 0            \\ \hline
		adv \(l_1\) keyboard        & 0              & 0.999         & 0.529         & 0.002        & 0.665         & 0            & 0.556         & 0            & 0.451         & 0            \\ \hline
		adv \(l_2\) keyboard        & 0              & 0.999         & 0.596         & 0.002        & 0.567         & 0.001        & 0.57          & 0            & 0.383         & 0            \\ \hline
		adv \(l_\infty \) keyboard      & 0              & 1             & 0             & \textbf{0.977}        & 0.139         & \textbf{0.107}        & 0.132         & \textbf{0.01}         & 0.267         & 0            \\ \hline \hline
		Org c-mouse            & 0.724          & 0             & 0.93          & 0            & 0.937         & 0            & 0.964         & 0            & 0.924         & 0            \\ \hline
		adv \(l_1\) c-mouse         & 0              & 0.999         & 0.518         & 0.004        & 0.911         & 0            & 0.914         & 0            & 0.884         & 0            \\ \hline
		adv \(l_2\) c-mouse         & 0              & 1             & 0.116         & 0.045        & 0.168         & 0.005        & 0.757         & 0            & 0.793         & 0            \\ \hline
		adv \(l_\infty \) c-mouse       & 0              & 0.999         & 0             & \textbf{0.995}         & 0             & \textbf{0.9}          & 0.014         & \textbf{0.013}        & 0.218         & 0            \\ \hline
	\end{tabular}
\end{table*}

\section{Conclusion}\label{conclusion}
In this study, we focus on investigating the connection between the search space of adversarial examples and the defense based on the frequency domain. Our empirical results demonstrate that the FGSM method in combination with \(l_\infty \)-norm produces the strongest adversarial examples. In this case, both the Gaussian and the Median filters are unable to restore identification to the ground truth label. However, when using \(l_\infty \)-norm to create adversarial examples, we also significantly reduce the quality of the original image compared to using \(l_1\) and \(l_2\) norm. In terms of similarities with the original image, \(l_1\) and \(l_2\) norm produce much better adversarial examples than \(l_\infty \) norm.

\section*{Acknowledgement}
We would like to thank Professor Akira Otsuka for the valuable suggestions on this research. This work was supported by the Iwasaki Tomomi Scholarship.
\bibliographystyle{IEEEtran}
\bibliography{JIACSA_Thang}

\end{document}